\documentclass[sn-mathphys-num]{sn-jnl}

\usepackage{graphicx}%
\usepackage{multirow}%
\usepackage{amsmath,amssymb,amsfonts}%
\usepackage{amsthm}%
\usepackage{mathrsfs}%
\usepackage[title]{appendix}%
\usepackage{xcolor}%
\usepackage{textcomp}%
\usepackage{manyfoot}%
\usepackage{booktabs}%
\usepackage{caption}
\usepackage{algorithm}%
\usepackage{algorithmicx}%
\usepackage{algpseudocode}%
\usepackage{listings}%
\usepackage{graphicx}
\usepackage{subfigure}
\usepackage{float}

\theoremstyle{thmstyleone}%
\theoremstyle{thmstyletwo}%

\theoremstyle{thmstylethree}%

\begin{document}

\title[Explainable AI-Powered Framework for Video-Based Skill Assessment in Cataract Surgery]{Explainable AI-Powered Framework for Video-Based Skill Assessment in Cataract Surgery}


\author[1]{\fnm{Mohammad Javad} \sur{Ahmadi}}\email{mjahmadi@email.kntu.ac.ir}

\author*[1]{\fnm{Hamid D.} \sur{Taghirad}}\email{taghirad@kntu.ac.ir}

\affil*[1]{\orgdiv{Applied Robotics and AI Solutions (ARAS), Faculties of Electrical and Computer Engineering}, \orgname{K.N. Toosi University of Technology}, \orgaddress{\city{Tehran}, \country{Iran}}}







\abstract{Persistent shortages in the surgical workforce and inherent limitations of traditional training methods highlight the necessity of automated, data-driven approaches in surgical education. This study addresses these challenges by introducing a novel, explainable AI-powered framework for automated skill assessment, specifically focusing on cataract surgery. We present the world's largest dataset of cataract surgery videos, comprising 2,000 recordings. Additionally, we propose an AI-powered analytical framework that employs advanced computer vision and signal-processing techniques to automatically evaluate surgical videos to derive objective, quantitative performance indicators that complement or potentially replace subjective scoring methods. A significant advantage of our framework over previous methods lies precisely in its explainability of outputs, elevating it beyond merely an opaque skill classification tool. Through experimental analysis of 83 cataract surgery videos, we demonstrate that the automatically computed metrics exhibit strong correlations with expert-based subjective evaluations, achieving up to 87\% accuracy in surgical skill assessment. Each metric was individually examined, and expert surgeons provided subjective ratings using the newly introduced Capsulorhexis Skill Assessment System (CSAS). These subjective assessments were compared with ten objective motion-based metrics extracted through our framework. The results indicated a robust correlation between subjective ratings and automated indicators, underscoring the framework's capacity to accurately model surgical expertise. 
}

\keywords{Surgical Education, Skill Assessment, Cataract Surgery, Computer Vision, Dataset, Video Analysis, Motion Analysis}



\maketitle

\section{Introduction}\label{sec1}

Global demand for surgical services is significantly increasing; however, access to these procedures remains uneven. Notably, low- and middle-income countries face a shortfall of 143 million essential surgeries each year \cite{vumc2024}. Expectations indicate that every country will require at least 5,000 surgical procedures per 100,000 people annually by 2030 \cite{UCSF_Surgery2030}. Addressing this gap necessitates a significant expansion in healthcare infrastructure and personnel, including doubling the surgical workforce over the next fifteen years \cite{UCSF_Surgery2030}. 

Nevertheless, current educational and assessment methods in surgical training, such as the Master-Apprentice Model (MAM), are inadequate to achieve these objectives. These conventional methods face several challenges, notably the reliance on subjective performance evaluations rather than objective, measurable standards, which often leads to biased and inconsistent assessments \cite{Augestad2019}. Furthermore, the high workload of trainers limits the time available for providing detailed and comprehensive feedback \cite{mbzuai2024ai_surgical_skill_optimization}. 

The growing gap between surgical needs and available resources, combined with the limited integration of advanced technology in training, has resulted in a lack of skilled surgeons, longer patient wait times, increased surgical errors, and patient deaths \cite{c8}. Research indicates that early-career surgeons are nearly nine times more likely to commit procedural errors compared to experienced surgeons \cite{Papaspyros2010}, underscoring the need for improved training and assessment methods \cite{Tevis_Kennedy_2013, Semel_Lipsitz_Funk_Bader_Weiser_Gawande_2012}. By implementing enhanced training programs, healthcare systems could prevent approximately 42,000 hospital readmissions annually, save up to \$3.62 million in costs \cite{c5}, and mitigate 67\% of cases leading to adverse events due to inadequate instrument handling skills \cite{c9}.

In this context, advancements in artificial intelligence (AI) present promising opportunities to enhance surgical procedures and transform training paradigms. One approach is the recording and analysis of surgical videos. Unlike sensor and kinematic data, which require expensive equipment, video recordings are readily available in most operating rooms through devices like surgical microscopes. Beyond their accessibility, surgical videos provide a comprehensive view of the entire process, capturing the surgeon's movements, the patient's condition, and the progress of the surgical workflow \cite{Gallant2021}.

Furthermore, integrating surgical recordings into training programs has become an invaluable resource for enhancing practical skills and academic research in the surgical field \cite{Levin2021, Theator2024, Gallant2021}. Novice surgeons report that reviewing videos and receiving feedback helps them prepare for surgeries, improve their anatomical knowledge, and refine their skills by analyzing past performances \cite{Theator2024}.

Despite the recognized benefits of surgical video materials, their widespread use remains limited due to challenges such as the time-consuming and expensive review of extensive video archives \cite{Cheikh2023, Awshah_Bowers_Eckel_Diab_Ganam_Sujka_Docimo_DuCoin_2024, Eckhoff_Rosman_Altieri_2023}. These findings highlight the need for advanced systems that streamline surgical video management, review, and analysis. Implementing intelligent platforms to create annotated video libraries could enhance training programs and improve clinical outcomes.

While current research efforts in automated surgical video analysis frequently focus on identifying operative phases and workflows, data-driven methodologies for objective video-based surgical skill assessment remain limited \cite{DTaghirad}. Current research predominantly utilizes deep neural network architectures to analyze surgical videos by processing them either as individual frames or as short video segments. These neural networks are designed to extract spatiotemporal features from video that correspond to distinct skill levels. 

Funke et al. \cite{Funke_Mees_Weitz_Speidel_2019} proposed an end‑to‑end deep‑learning pipeline that classifies surgical skills from short video clips using an inflated 3‑D CNN and a Temporal Segment Network. Their evaluation was conducted on JIGSAWS \cite{gao2014jigsaws}, a bench‑top simulation dataset, not footage from the real operations. Although the reported accuracies exceeded 95\%, this performance was achieved only for a three-class classification. Moreover, the reliance on simulated data and the absence of interpretable feedback constrain the approach’s immediate clinical relevance.

Kitaguchi et al. \cite{Kitaguchi_Takeshita_Matsuzaki_Igaki_Hasegawa_Ito_2021} trained a 3‑D CNN on 40-second clips from 74 videos of laparoscopic colorectal procedures labeled with Endoscopic Surgical Skill Qualification System (ESSQS) scores. Their model achieved a mean accuracy of 75\% when classifying surgical performance into low, medium, or high skill levels. However, the moderate accuracy, use of a small and selectively sampled dataset, and reliance on a black-box three-tier output limit the method’s robustness and its utility for providing meaningful, detailed educational feedback.

Zia et al. \cite{Zia_Guo_Zhou_Essa_Jarc_2019} proposed RP‑Net‑V2, a CNN–LSTM architecture that segments robotic‑assisted radical prostatectomies into 12 steps and evaluates performance through task‑based efficiency metrics, achieving a Jaccard index of 0.85 and strong correlations with expert scores. However, the model’s reliance on raw RGB frames leads to frequent confusion between visually similar tasks, the class imbalance is left unaddressed, a simple running‑window median filter shifts some predicted boundaries by over a minute, and the ground‑truth labels originate from a single annotator.

Lajkó et al. \cite{Lajko_NagyneElek_Haidegger_2021} proposed a 2D video-based skill assessment method using sparse optical flow and standard classifiers on the JIGSAWS dataset, achieving around 80\% accuracy. However, performance remained below kinematics-based methods, and the approach relied on manually selected tool regions, limiting automation.

Although deep learning models have achieved promising results in medical applications, their widespread use in sensitive clinical settings remains constrained by several issues, including the black box nature of most neural networks, the demand for extensive expert-annotated datasets, and limited capacity for delivering detailed, actionable feedback to surgeons. 

To address these challenges, this paper introduces an AI-powered framework that integrates explainable methods for objective and automated skill evaluation in surgical videos. Although the framework is applicable to a wide range of surgical procedures, its effectiveness should be independently evaluated in each context. As a proof of concept, we focus on cataract surgery, specifically the capsulorhexis phase, as our primary case study. This choice is motivated by the fact that cataract surgery is among the most frequently performed surgeries worldwide \cite{technavio2023}.

Cataract remains the primary cause of preventable vision loss and significant visual impairment, particularly in developing nations \cite{Flaxman2017tw, Khairallah2015dc}. Phacoemulsification is the globally preferred surgery for cataract removal, making it a vital component of the residency educational curriculum for developing surgical competency \cite{Lacmanovic_Loncar2016ai}. Surgical competency requires deep knowledge, dexterity, micro-surgical skills, and dedication \cite{Neufeld2018gz}.

Creating a circular opening in the anterior surface of the cataracted lens, called capsulorhexis, is a fateful step in phacoemulsification cataract surgery. It's a significant challenge for first-year residents \cite{Bharucha2020in, Prakash2009ed}. Trainee surgeons require years of training and evaluation, with evolving methods and tools for learning and assessment. Rubrics, with their structured approach, are the conventional tools in skill assessments. 

Several rubrics are available for evaluating cataract surgery, such as the International Council of Ophthalmology’s Ophthalmology Surgical Competency Assessment Rubric (ICO-OSCAR) \cite{Golnik2011db}, Objective Assessment of Skills in Intraocular Surgery (OASIS) \cite{Cremers2005dp}, Global Rating Assessment of Skills in Intraocular Surgery (GRASIS) \cite{Cremers2005aj}, and Objective Structured Assessment of Cataract Surgical Skill (OSACSS) \cite{Saleh2007bo}. However, these rubrics typically require direct trainer supervision, leading to high costs, time-consuming processes, and potential biases.

Our proposed framework overcomes these limitations by automating the assessment process, eliminating the need for direct trainer supervision. This reduces both costs and time, while minimizing the potential for biases, providing a more efficient and objective method for evaluating surgical skills.

In the following sections, we present our newly developed cataract surgery video dataset, the largest in the world in terms of the number of data, and comprehensive annotations. We then describe our method for intelligent, automated, and objective skill evaluation, detailing the framework’s architecture and operational steps. In the Results section, we report the performance of the proposed framework on 83 cataract surgery videos, accompanied by extensive analyses that underscore both the significance of this dataset and the method’s effectiveness. We also discuss each objective metric in depth to illuminate various dimensions of skill assessment. Finally, in the Conclusion, we summarize the key findings and outline potential directions for future research.

\section{Methods}\label{sec2}
In this section, we present a large-scale surgical video dataset designed to support interdisciplinary AI research and serve as ground truth for validating our automated skill assessment systems. We then introduce a novel AI-powered framework that seamlessly processes these videos, generating interpretable metrics of surgical performance and advancing the reliability of automated skill evaluation in clinical practice.

\subsection{Dataset Crafting Description}\label{sec2sub1}

Cataract surgery is performed with precise instruments under a surgical microscope, while an attached camera captures the operation from the surgeon’s perspective. The ARAS Group and Farabi Eye Hospital jointly developed an infrastructure for the systematic acquisition of these surgical videos. 

From 2021 to 2024, more than 2,500 videos were recorded at up to 60 frames per second with a resolution of 1920$\times$1080 pixels. The video selection process involved rigorous exclusion criteria that filtered out videos with technical flaws such as substandard image quality, and low resolution.

Having established a high-quality dataset, the next step is to select an appropriate method for skill-based annotation. In the domain of surgical video analysis, two principal methods have been utilized to annotate surgeon skill. The first method involves identifying individual surgeons and estimating their proficiency based on metrics such as academic rank and accumulated surgical experience, as utilized in the Cataract-101 dataset \cite{Cataract101}.

In contrast, the second approach relies on expert review of surgical videos, where evaluators assign skill scores according to predefined standards. This expert-based approach generally provides a more accurate and unbiased evaluation of surgical performance than identity-based methods.

This study introduces a video-based skill annotation method that segments surgical procedures into distinct phases, each evaluated with specialized criteria. In our case study, video segments corresponding to the capsulorhexis phase were extracted from the recorded procedures, and detailed annotations were assigned to these segments using a new system developed to standardize the review process of capsulorhexis videos.

The proposed assessment framework was validated through a comprehensive review of existing surgical skill assessment systems \cite{d16, d17, d18, d19, Golnik_2011}, carried out in collaboration with two experienced surgeons. Based on this review, we introduced the Capsulorhexis Skill Assessment System (CSAS). \autoref{tab:csas} details the 11 indicators of CSAS, incorporating seven subjective and four objective metrics: rhexis position, shape, size, and time. Notably, the indicators ``Commencement of Flap" and ``Circular Completion" follow the ICO-OSCAR \cite{Golnik_2011} guidelines, whereas ``Tissue Handling", ``Instrument Handling", and ``Microscope Use" have been adapted from the GRASIS \cite{Cremers_2005} scoring system.

In this study, 83 preprocessed surgical videos have been independently reviewed and rated by three trained medical professionals using CSAS subjective indicators. This review process resulted in crafting the cataract surgical video skill assessment dataset (CVSAD-83) \cite{lmm}. Each indicator was rated on a five-point scale, where higher scores indicate greater proficiency. The reliability of these annotations was confirmed by an intraclass correlation coefficient (ICC) of 0.79 and a Cronbach’s alpha coefficient of 0.93, indicating high inter-rater reliability and internal consistency. To facilitate the categorization of surgeon skills according to the Dreyfus Model \cite{dreyfus_skill_levels}, an overall score (OS) was computed for each video by averaging the scores of all indicators. This OS, ranging from 1 to 5, is a reliable metric for skill classification.

\begin{sidewaystable}[htbp]
\centering
\scalebox{0.9}{
\renewcommand{\arraystretch}{1.2} 
\begin{tabular}{p{3.2cm} p{4.8cm} c p{4.8cm} c p{4.8cm} c}
\hline
\textbf{Indicator} & \multicolumn{2}{c}{\textbf{Novice}} & \multicolumn{2}{c}{\textbf{Advanced Beginner}} & \multicolumn{2}{c}{\textbf{Competent}} \\
\hline
 & \textbf{Description} & \textbf{Score} & \textbf{Description} & \textbf{Score} & \textbf{Description} & \textbf{Score} \\
\hline

\textbf{Instrument Handling} 
 & Repeated, Abrupt, Awkward/Bizarre, and Harsh Movements; Endless Insertion/Entry and Exit 
 & 1--2 
 & Selected/Occasional Inappropriate Movements 
 & 3--4 
 & Fine and Smooth Movements; No Inappropriate Movements 
 & 5 \\
\hline

\textbf{Motion}
 & Unsure Surgical Plan; Needless in-doubt Movements
 & 1--2
 & Certain Surgical Plan; Occasional/selected Unnecessary Movements
 & 3--4
 & Maximum Effective Movements; No Unnecessary Movements
 & 5 \\
\hline

\textbf{Commencement of flap}
 & Tentative Chases rather than Controlled; Numerous disruptions in the cortex
 & 1--2
 & Flap Pull-up After 2--3 Tries; Subtle Cortex Disruptions
 & 3--4
 & Delicate and Controlled Approach; No Cortex disruption
 & 5 \\
\hline

\textbf{Tissue Handling}
 & Using Unnecessary Force; Damage to Cornea and Conjunctiva
 & 1--2
 & Suitable Tissue Interactions; Unintentional/Accidental Tissue Damage
 & 3--4
 & Excellent Tissue Interactions; No Tissue Damage
 & 5 \\
\hline

\textbf{Microscope Use}
 & Multiple Recentering; Multiple Refocus
 & 1--2
 & Few attempts to recenter/refocus
 & 3--4
 & Keeps Eye Centered; Good Focused View
 & 4--5 \\
\hline

\textbf{Circular Completion}
 & Not Able to Achieve Circular Rhexis; Entrance to Dangerous Regions (Extend to the Periphery)
 & 1--2
 & Difficulty Achieving Rhexis
 & 3--4
 & Rapid, Unaided Completion of Rhexis
 & 5 \\
\hline

\textbf{Adverse Events}
 & Rhexis Tear; Entering Hazardous Areas; Unable to Recognize Adverse events. Inappropriate Reaction
 & \checkmark
 & --
 & --
 & --
 & -- \\
\hline

\textbf{Position of Rhexis}
 & $>$\,1\,mm
 & 1--2
 & 0.4--1\,mm
 & 3--4
 & $<$\,0.4\,mm
 & 5 \\
\hline

\textbf{Shape of Rhexis}
 & Irregular Shape and Margins
 & 1--2
 & Semi-Circular Shape; Minor Irregularities along Edges
 & 3--4
 & Perfectly Circular
 & 4--5 \\
\hline

\textbf{Size of Rhexis}
 & $<$\,4.25\,mm or $>$\,6.25\,mm
 & 1--2
 & 4.25--4.75\,mm or 5.75--6.25\,mm
 & 3--4
 & 4.75--5.75\,mm (5--5.5\,$\pm$\,0.25\,mm)
 & 5 \\
\hline

\textbf{Time}
 & $>$\,75\,seconds
 & 1--3
 & --
 & --
 & $\leq$75\,seconds
 & 4--5 \\
\hline
\end{tabular}
} 
\caption{Capsulorhexis Skill Assessment System (CSAS)}
\label{tab:csas}
\end{sidewaystable}

\subsection{AI-Powered Objective Skill Assessment Framework}
This paper introduces an innovative framework designed to enhance the explainability of AI-powered systems for automatic surgical skill assessment through video analysis. The framework begins with a computer vision algorithm that tracks and segments surgical tools and relevant tissues, extracting pixel-level motion information related to these elements. Subsequently, this raw pixel data is processed through a post-processing stage, converting it into actionable information for surgical skill assessment. Finally, an objective evaluation system is developed, which evaluates the surgeon’s skills based on motion data and formulated criteria extracted from the logic of subjective indices. 

\subsubsection{Motion Data Extraction and Processing} \label{ewd333}

In this section, we present a detailed overview of the methods used to extract and process video-based motion data. To accomplish this, our proposed framework integrates state-of-the-art computer vision algorithms for segmentation and tracking, such as SAM-2 \cite{ravi2024sam2segmentimages} or YOLO-11 \cite{khanam2024yolov11overviewkeyarchitectural}.

In this study, pre-trained models developed in our previous studies \cite{Ahmadi2021, Lafouti2022, Gandomi2023} were utilized to segment the capsulorhexis instrument and the corneal tissue in the videos of CVSAD-83 dataset. The motion data obtained through tracking these components provides the essential data to evaluate surgical skills.

Following the segmentation-tracking stage, the frame-by-frame coordinates of the capsulorhexis tool-tip and the corneal center are identified, forming the fundamental motion data for analysis. However, to ensure the resulting measurements are meaningful for skill evaluation, a proper referencing system must be established. Without appropriate referencing, raw positional measurements may lead to inaccurate conclusions about skill performance.

Absolute referencing, in which each component’s position is determined relative to a fixed point in the frame (e.g., the top-left corner), can be misleading. Rapid shifts in a tool’s location with respect to this fixed point may arise from patient eye movements or changes in the microscope’s field of view, rather than a lack of surgical precision. 

To address this limitation, a relative referencing scheme provides a more reliable measure for skill-related analysis by representing the tool-tip coordinates in relation to the target tissue (in this case, the corneal center). Transitioning to a relative reference system effectively translates sudden absolute movements into smoother, more interpretable variations in the tool-tip’s position, allowing for a more accurate assessment of surgical performance.

After extracting these relative positions, the data must be scaled to account for varying magnifications in the surgical microscope across different videos. One practical method is to measure the corneal diameter in pixels using computer vision techniques and then convert these measurements to physical units based on the known corneal diameter (11.71 ± 0.42 mm \cite{Rfer2005}). Although this scaled data may not precisely replicate real-world dimensions, it serves its primary purpose of capturing the essential motion trends indicative of surgical proficiency. 

Finally, higher-order motion features such as velocity, acceleration, and jerk are derived from the scaled relative positions for objective skill evaluation in subsequent analytical steps. By focusing on the dynamics of tool movement rather than on absolute positional data, this approach provides a more reliable framework for subsequent analytic processes to distinguish between different levels of surgical expertise.

\subsubsection{Objective Metrics for Automated Surgical Skill Assessment}

Following the discussion on motion data extraction and processing methods, this section presents the objective metrics for automated surgical skill assessment. These metrics were derived through expert consultation and by adapting the principles from subjective assessment systems.

\begin{enumerate}
    \item 
Procedure Time (PT): The total time for completing the surgery, \( T \), is computed by subtracting the start time from the end time, as in \autoref{eq101}: 

\begin{equation} \label{eq101}
    \text{PT} = t_{\text{end}} - t_{\text{start}}.
\end{equation}

    \item 
Path Lenrth (PL): This metric is computed by summing the magnitudes of the tool's relative positional coordinates. Specifically, if \( (x_i, y_i) \) represents the position at the \( i \)-th time instant, then the path length is given by \autoref{eq102}:

\begin{equation} \label{eq102}
    \text{PL} = \sum_{i=1}^{n} \sqrt{x_i^2 + y_i^2},
\end{equation}

where \( n \) is the total number of sampled points in the recorded trajectory. 
 
    \item 
Number of Alternating (Back-and-Forth) Movements (NAM): This metric reflects the count of tool direction changes. A change in the sign of the velocity along any axis indicates a reversal in tool motion. The total number of such movements is calculated as in \autoref{eq103}:

\begin{equation} \label{eq103}
    \text{NAM} = \sum_{i=1}^{n-1} \delta (\text{sign}(v_{i+1}) - \text{sign}(v_i)),
\end{equation}

where \( \text{sign}(\cdot) \) extracts the velocity direction along either \( x \) or \( y \) axis, and \( \delta(\cdot) \) is 1 if a sign change is detected, otherwise 0.

    \item 
Motion Difficulty (MD): Abrupt changes in acceleration indicate higher jerk, which are associated with reduced smoothness \cite{Hwang2005}. The overall difficulty is quantified by summing the magnitude of the jerk vector, as given by \autoref{eq104}:

\begin{equation} \label{eq104}
    \text{MD} = \sum_{i=1}^{n} \sqrt{j_{x_i}^2 + j_{y_i}^2},
\end{equation}

where \( j_{x_i} \) and \( j_{y_i} \) are the jerk components along the \( x \) and \( y \) axes, respectively. 

    \item 
Energy Consumption (EC): By assuming the needle mass is negligible and constant, the kinetic energy concept \( \frac{1}{2} m v^2 \) can be used to estimate the total energy. Let \( (v_{x_i}, v_{y_i}) \) be the velocity components at time \( i \). The \text{EC} is estimated using \autoref{eq105}:

\begin{equation} \label{eq105}
    \text{EC} = \sum_{i=1}^{n} \frac{1}{2} (v_{x_i}^2 + v_{y_i}^2).
\end{equation}

    \item 
Number of Local Peaks (NLP): The local peaks in speed, acceleration, or jerk profiles are counted using computational methods such as the \texttt{find\_peaks} Python function in \texttt{scipy.signal}. Let NLP denote the total number of detected peaks across these profiles, formalized as in \autoref{eq106}:

\begin{equation} \label{eq106}
    \text{NLP} = \text{Peaks}(\text{speed}) + \text{Peaks}(\text{acceleration}) + \text{Peaks}(\text{jerk}),
\end{equation}

where \( \text{Peaks}(\cdot) \) returns the count of local peaks in the corresponding signal.

    \item 
Total Force (TF): Inspired by the physical relation \( \text{Force} = m \times a \) and assuming negligible needle mass, the total force is estimated by summing the magnitude of the acceleration vector. \autoref{eq107} defines Total Force:

\begin{equation} \label{eq107}
    \text{TF} = \sum_{i=1}^{n} \sqrt{a_{x_i}^2 + a_{y_i}^2}.
\end{equation}

    \item 
Tissue Interaction (TI): Sharp deviations from the mean acceleration can indicate sudden forces that might harm tissue. By comparing local acceleration peaks with the mean, an interaction index (TI) is computed as indicated in \autoref{eq108}:

\begin{equation} \label{eq108}
    \text{TI} = \sum_{i=1}^{n} \left| \sqrt{a_{x_i}^2 + a_{y_i}^2} - \mu_a \right|,
\end{equation}

where \( \mu_a \) is the average magnitude of acceleration. 

    \item 
Speed Frequency Smoothness (SFS): The mean frequency magnitude in the fast Fourier transform (FFT) of the speed signal is used to indicate frequency smoothness. \autoref{eq109} details the computation:

\begin{equation} \label{eq109}
    \text{SFS} = \frac{1}{n} \sum_{i=1}^{n} \left| \text{FFT}(v_i) \right|,
\end{equation}

where \( \text{FFT}(\cdot) \) denotes the FFT of the velocity at each sampled interval. 

    \item 
Spectral Entropy (SE): The distribution of the signal’s energy across frequencies is captured by the spectral entropy, described in \autoref{eq110}:

\begin{equation}\label{eq110}
   \text{SE} = - \sum_{f} P(f) \log P(f),
\end{equation}

where
$P(f) = \frac{|\text{FFT}(x_f)|^2}{\sum_{j} |\text{FFT}(x_j)|^2}$
is the normalized power spectral density of the signal.
    
\end{enumerate}

\begin{figure}[htbp]
    \centering
    \includegraphics[width=1\linewidth]{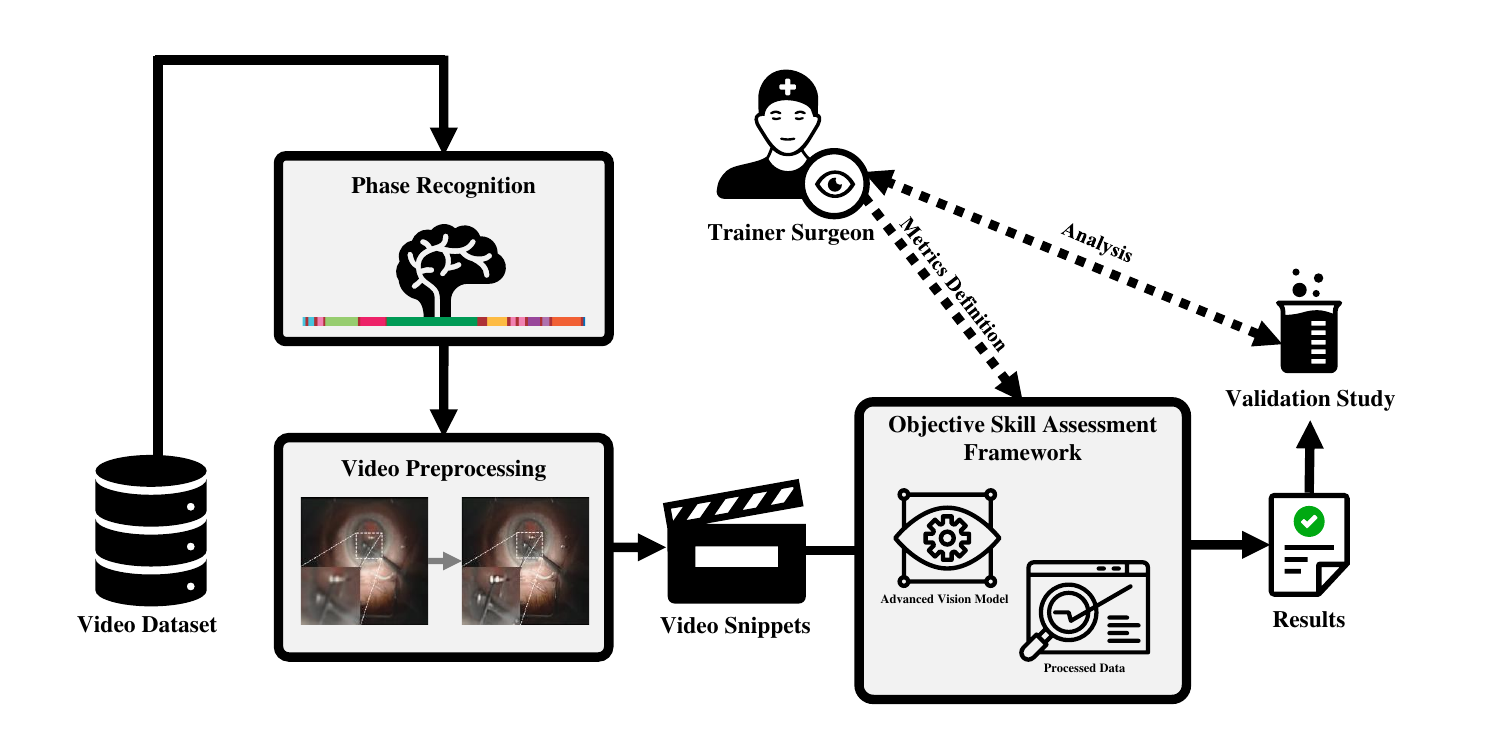}
    \caption{Overview of the proposed framework.}
    \label{fi555label}
\end{figure}

In the subsequent section, it will be demonstrated how these metrics, which were developed from subjective indices and motion data extracted from surgical videos, correlate with expert surgeons’ subjective assessments in the CVSAD-83 dataset. An overview of our proposed framework is presented in \autoref{fi555label}.

\section{Results}\label{sec2}

This section presents and discusses the findings of three primary analyses conducted on the CVSAD-83 dataset. First, we examine the distribution of subjective skill scores, offering insights into typical performance levels and areas for targeted improvement. Next, we assess the correlation between ten automated computed objective metrics and the overall subjective score (OS) assigned by expert surgeons, thereby evaluating the reliability of automated indicators as proxies for clinical expertise. Finally, we explore each metric’s discriminative power by clustering surgeons into different skill groups, comparing the results with ground-truth expert labels.

\begin{figure}[htbp]
    \centering
    \includegraphics[width=\linewidth]{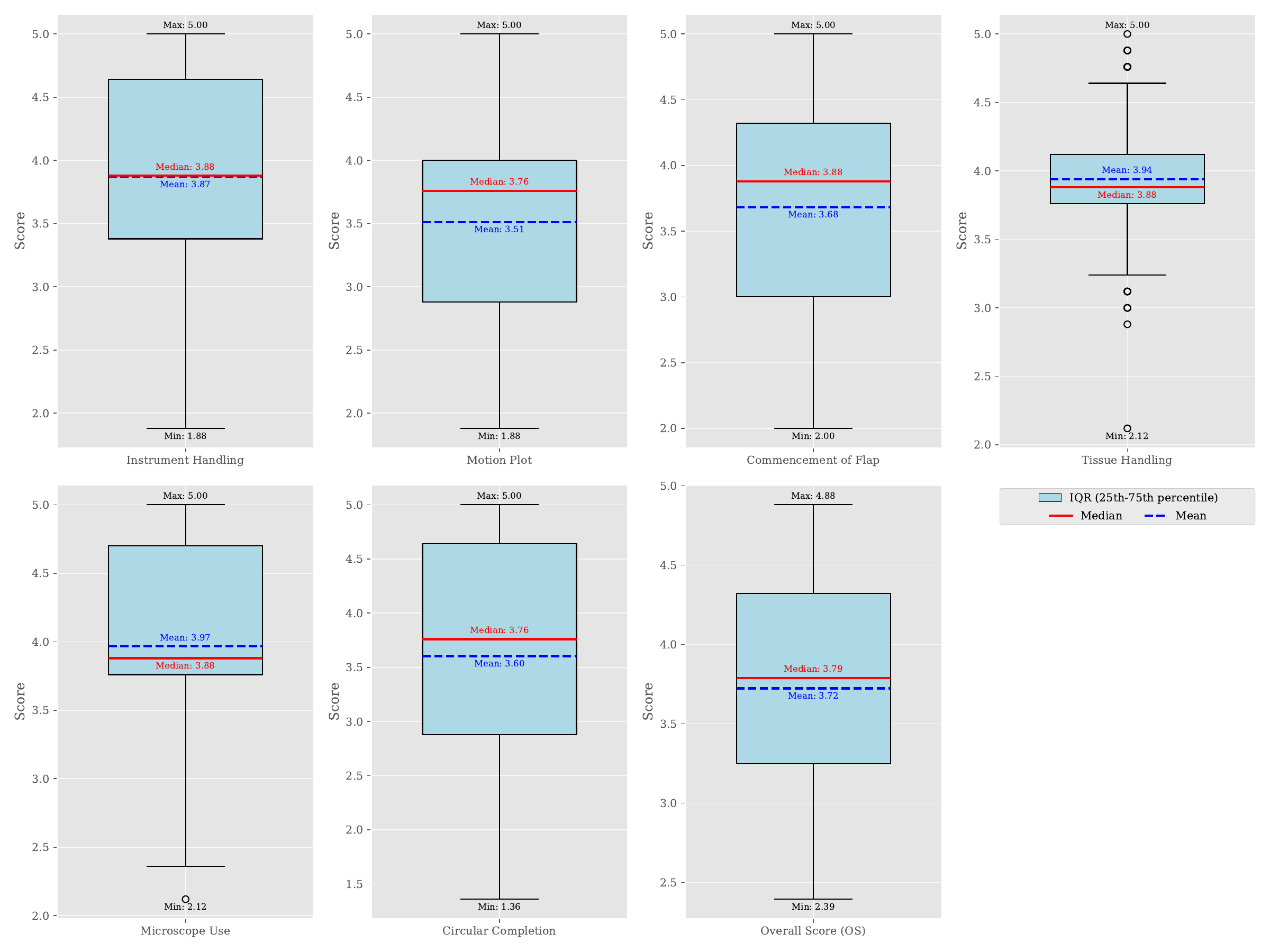}
    \caption{Overview of CSAS scores and overall performance from 83 CVSAD cases.}
    \label{fig569label}
\end{figure}

The box plots in \autoref{fig569label} illustrate the distribution of scores for the subjective indicators of the CSAS evaluation system, which were used to evaluate the 83 videos from the CVSAD dataset, in addition to the overall score (OS). In each plot, the box spans the interquartile range (IQR), with the median and mean values denoted by the horizontal lines and whiskers marking the full data range. Notably, most indicators exhibit median scores near or above 3.75, indicating that the majority of assessed performances cluster within the upper half of the 1–5 scale.

This insight into the distribution of performances helps trainers develop targeted interventions that can systematically improve surgical outcomes by addressing specific weak points discovered through the CSAS framework. For instance, the scores for Microscope Use and Tissue Handling appear slightly higher than those for Motion and Circular Completion, indicating that surgeons generally excel in maintaining focus and handling ocular tissues while having more varied performance on tasks requiring precise movement control. 

Following the subjective score distribution analysis, we explored the relationship between ten objective metrics generated by our automated assessment framework and the expert-assigned Overall Scores (OS). Establishing strong correlations between objective indicators and subjective assessments is crucial for validating the reliability of our framework.

\autoref{figcorrelation_plots} presents scatter plots for each metric against OS, with the horizontal axis representing the expert-assigned score (OS) value and the vertical axis showing the corresponding objective metric output. Data points are labeled by their video IDs, and a regression line highlights the general trend in each plot. A consistent inverse relationship emerges across all ten objective indicators, reflecting that surgeons assigned higher OS values typically exhibit lower measured values for objective metrics. A detailed and thorough analysis of the charts corresponding to each of the 10 objective indicators is presented as follows:

\begin{figure}[htbp]
    \centering
    \subfigure[Procedure Time (PT)]{
        \includegraphics[width=0.45\textwidth]{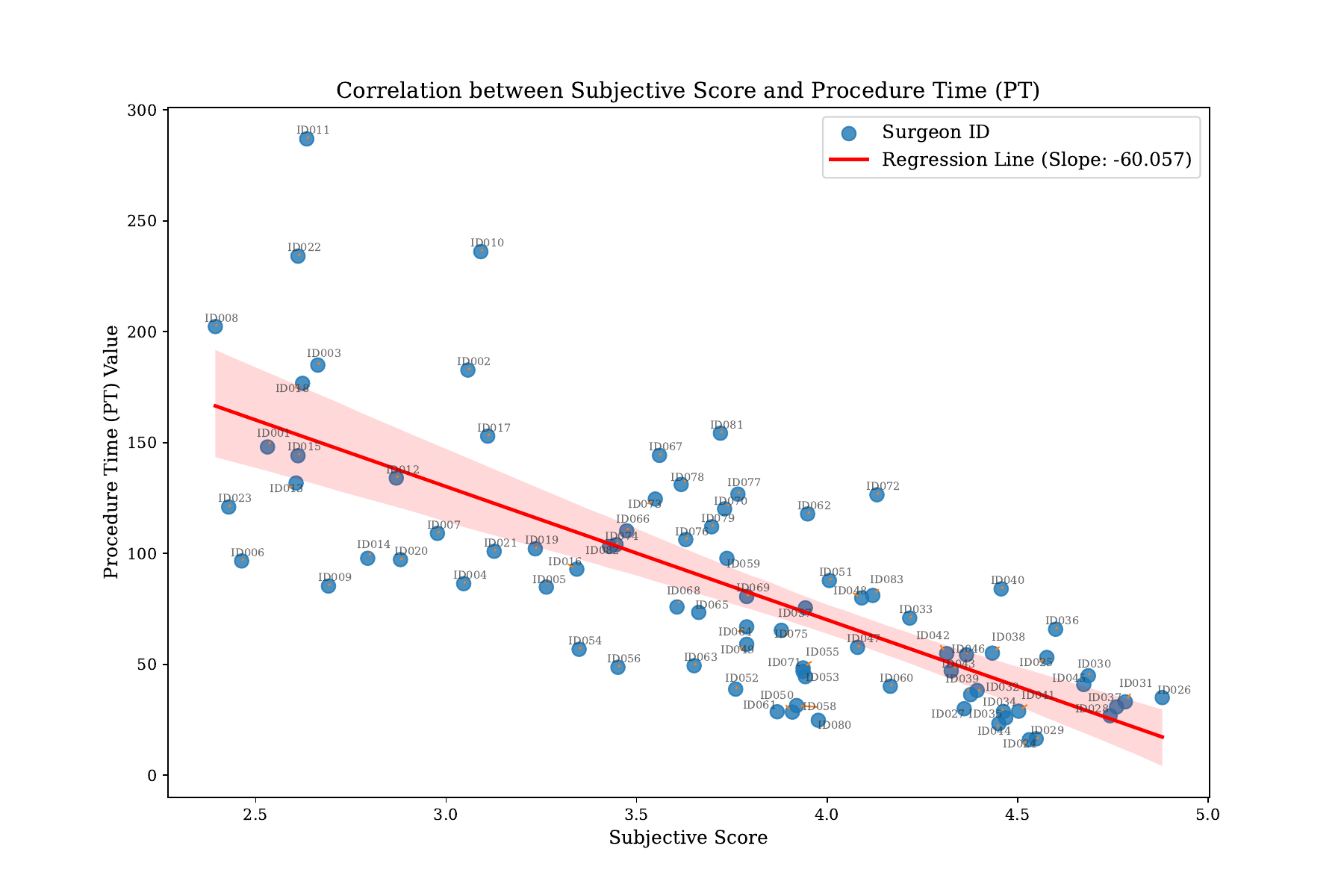} \label{PT}
    }
    \subfigure[Path Length (PL)]{
        \includegraphics[width=0.45\textwidth]{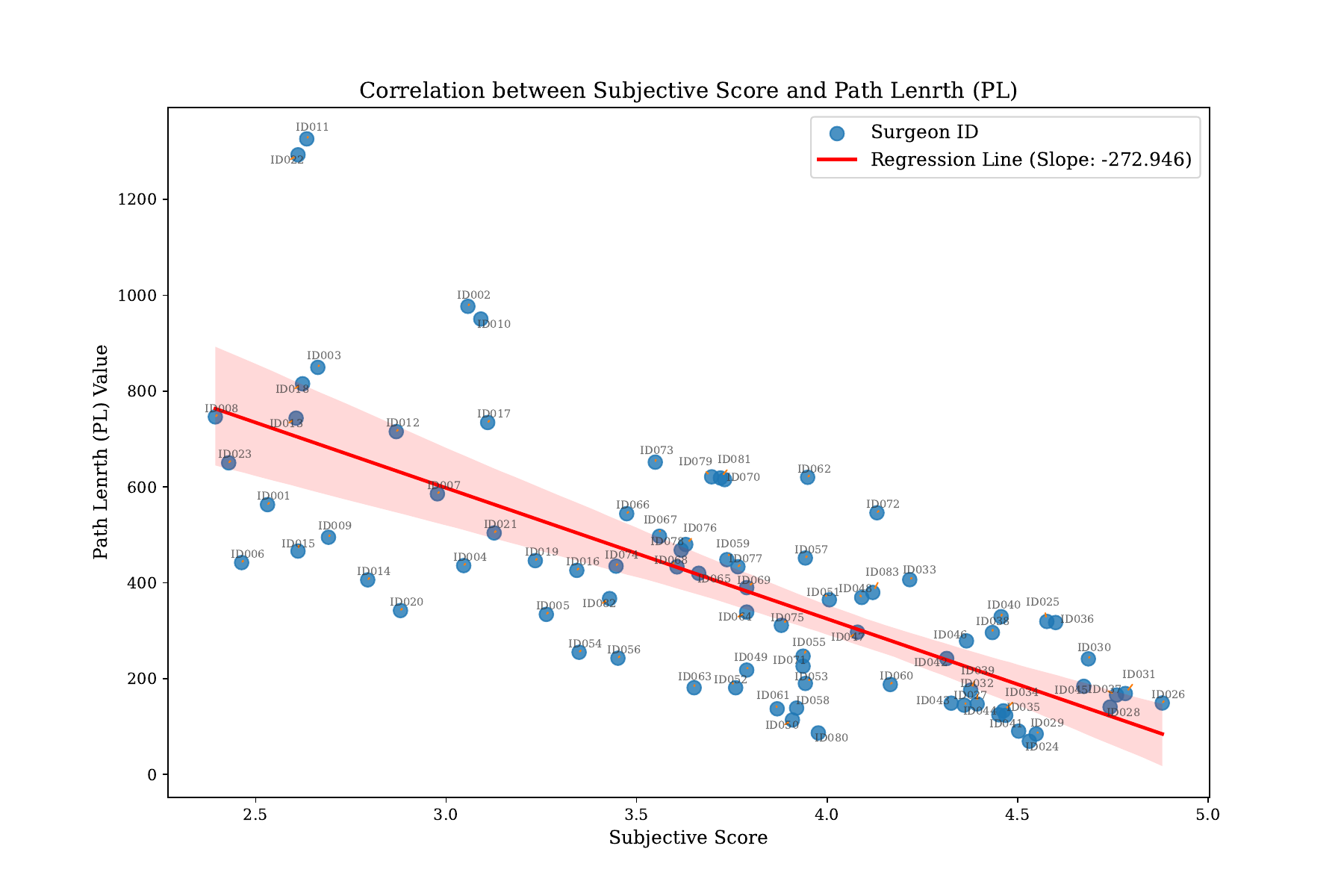}\label{PL}
    }
    
    \subfigure[Number of Alternating Movements (NAM)]{
        \includegraphics[width=0.45\textwidth]{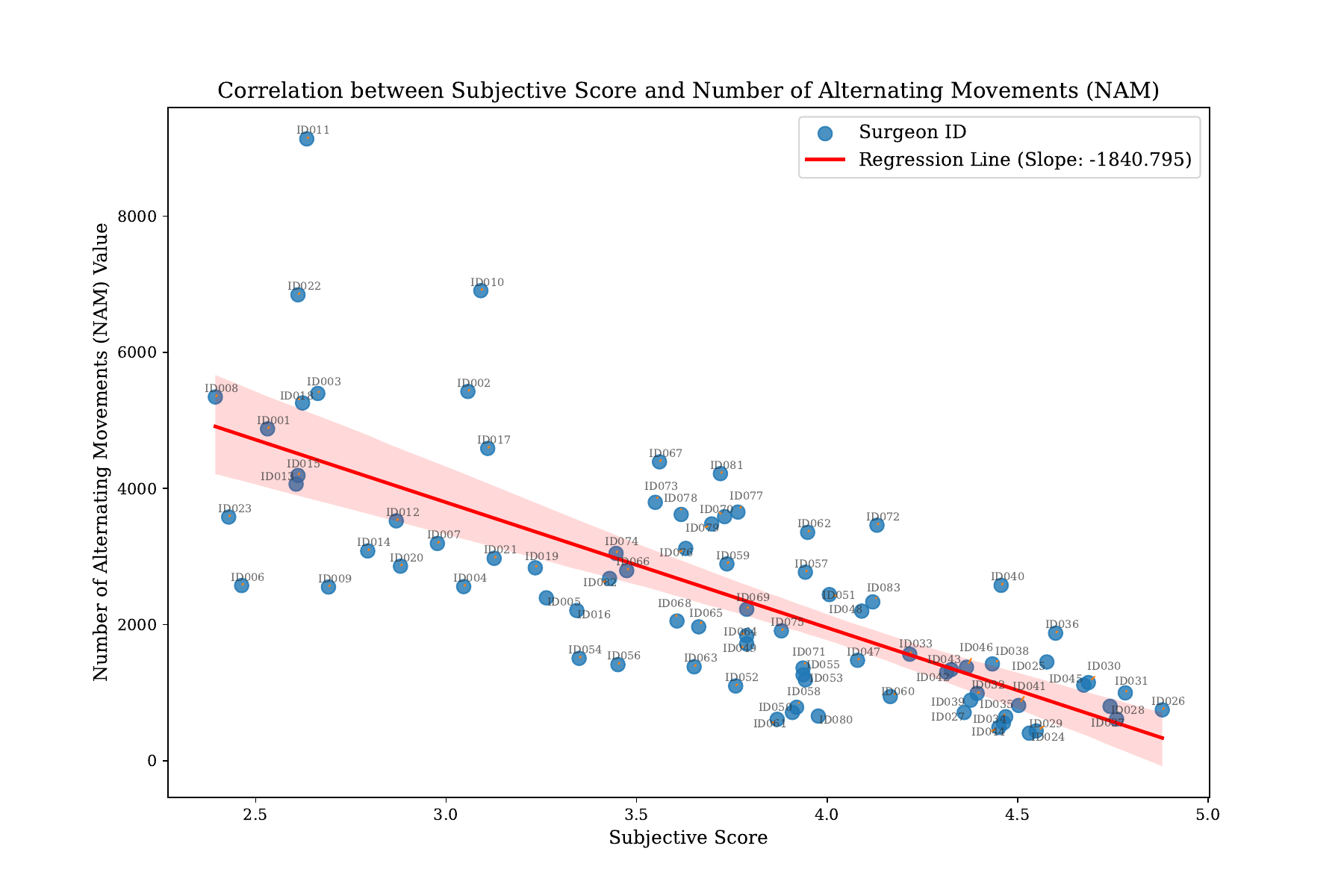}\label{NAM}
    }
    \subfigure[Motion Difficulty (MD)]{
        \includegraphics[width=0.45\textwidth]{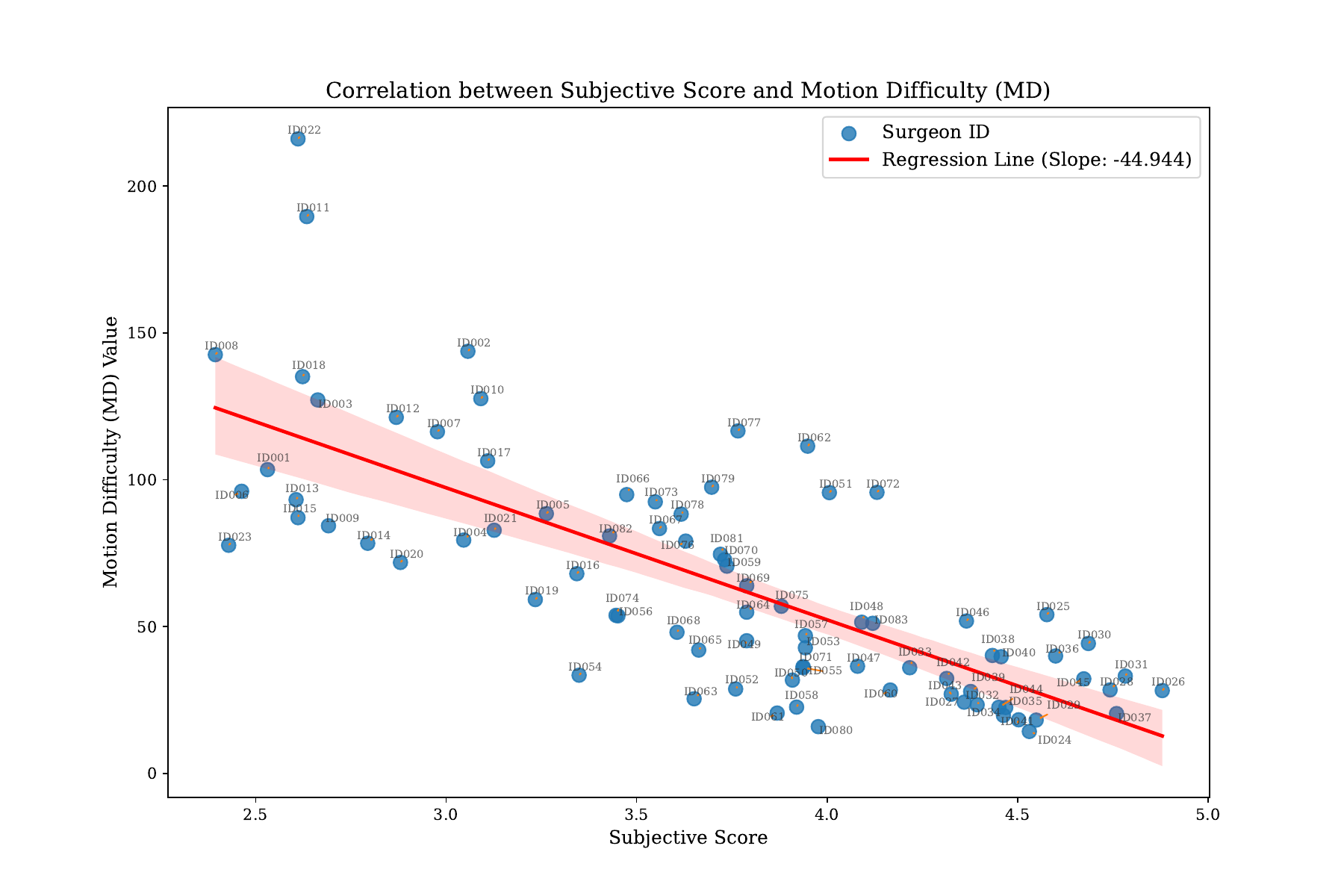}\label{MD}
    }
    
    \subfigure[Energy Consumption (EC)]{
        \includegraphics[width=0.45\textwidth]{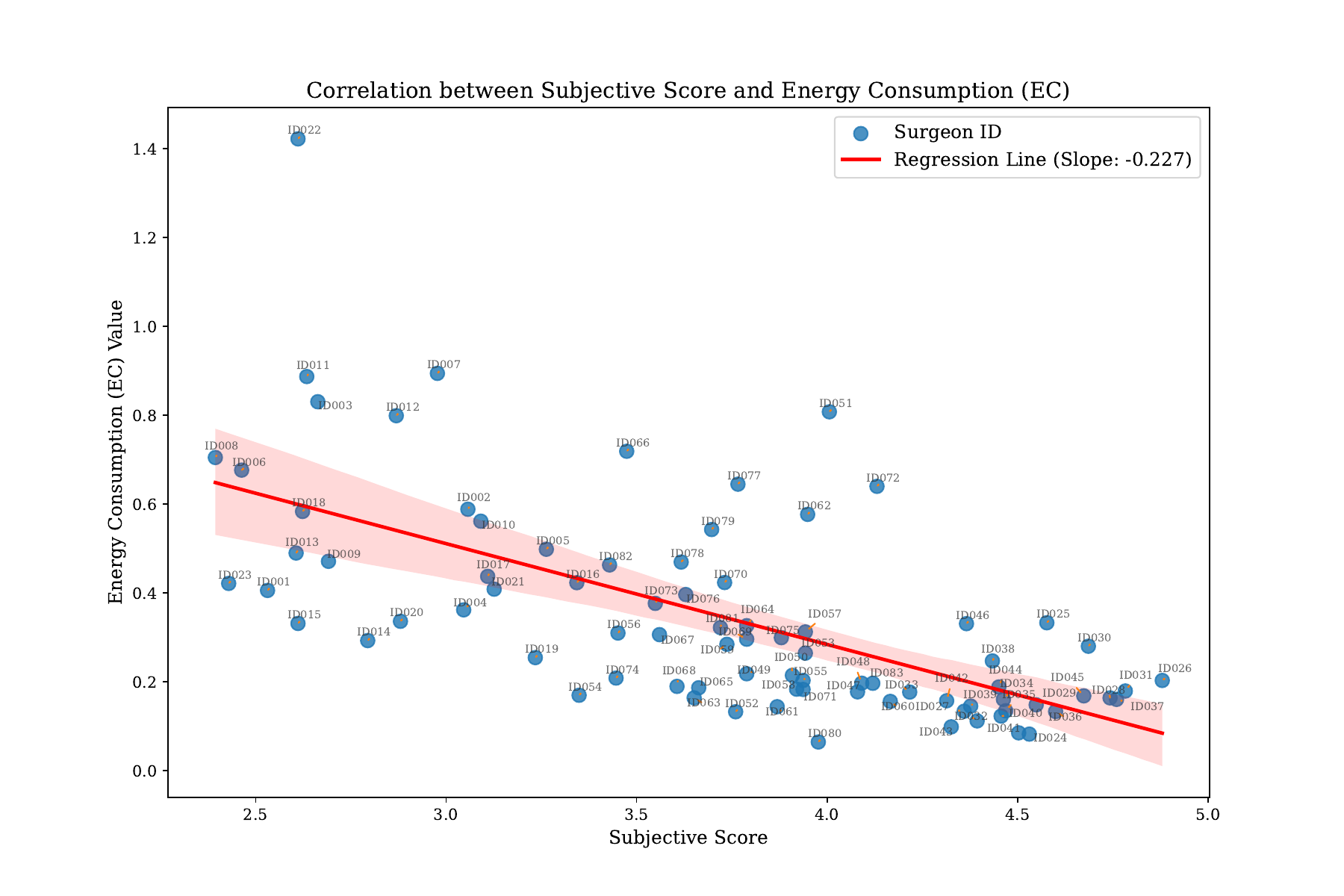}\label{EC}
    }
    \subfigure[Number of Local Peaks (NLP)]{
        \includegraphics[width=0.45\textwidth]{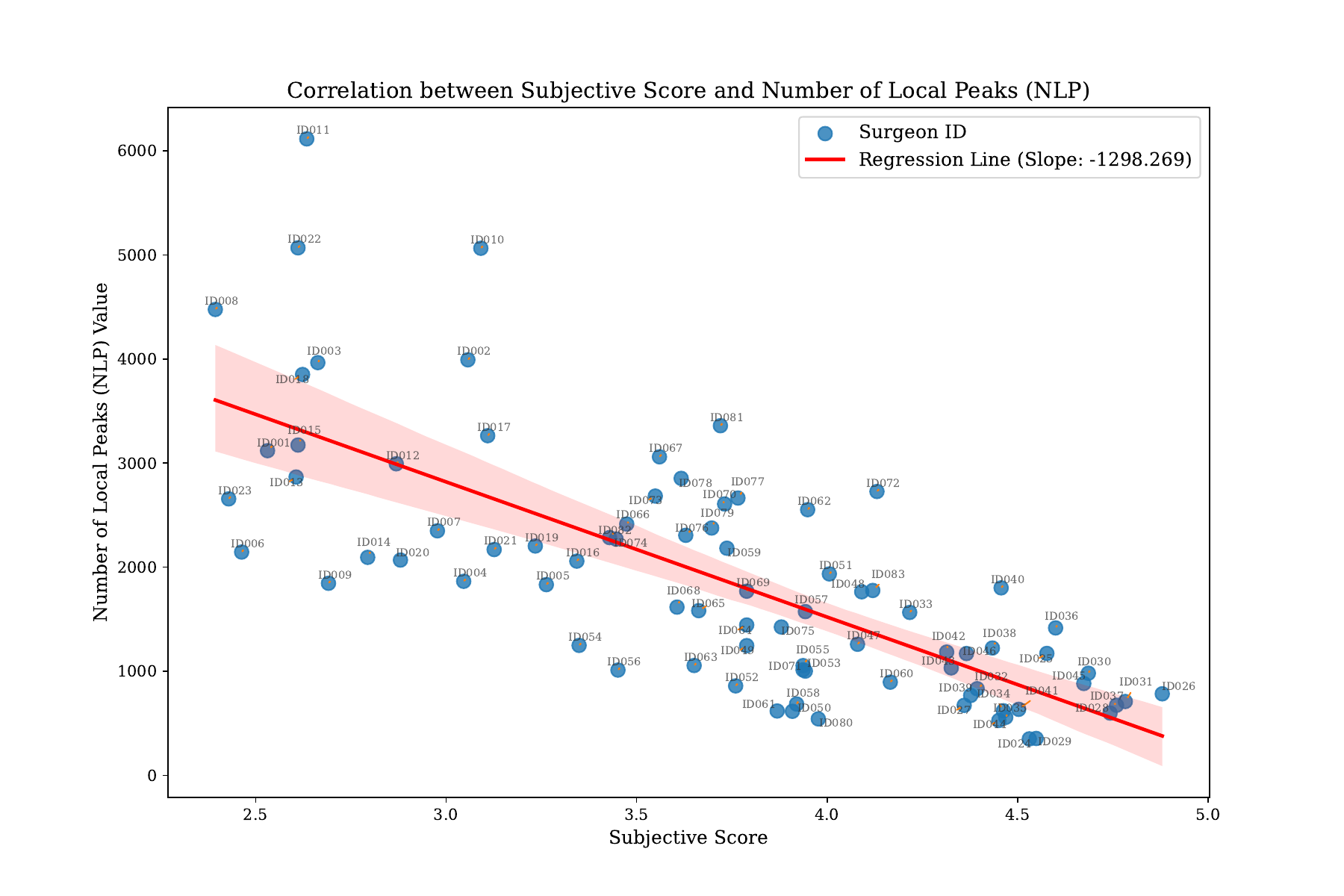}\label{NLP}
    }

    \subfigure[Total Force (TF)]{
        \includegraphics[width=0.45\textwidth]{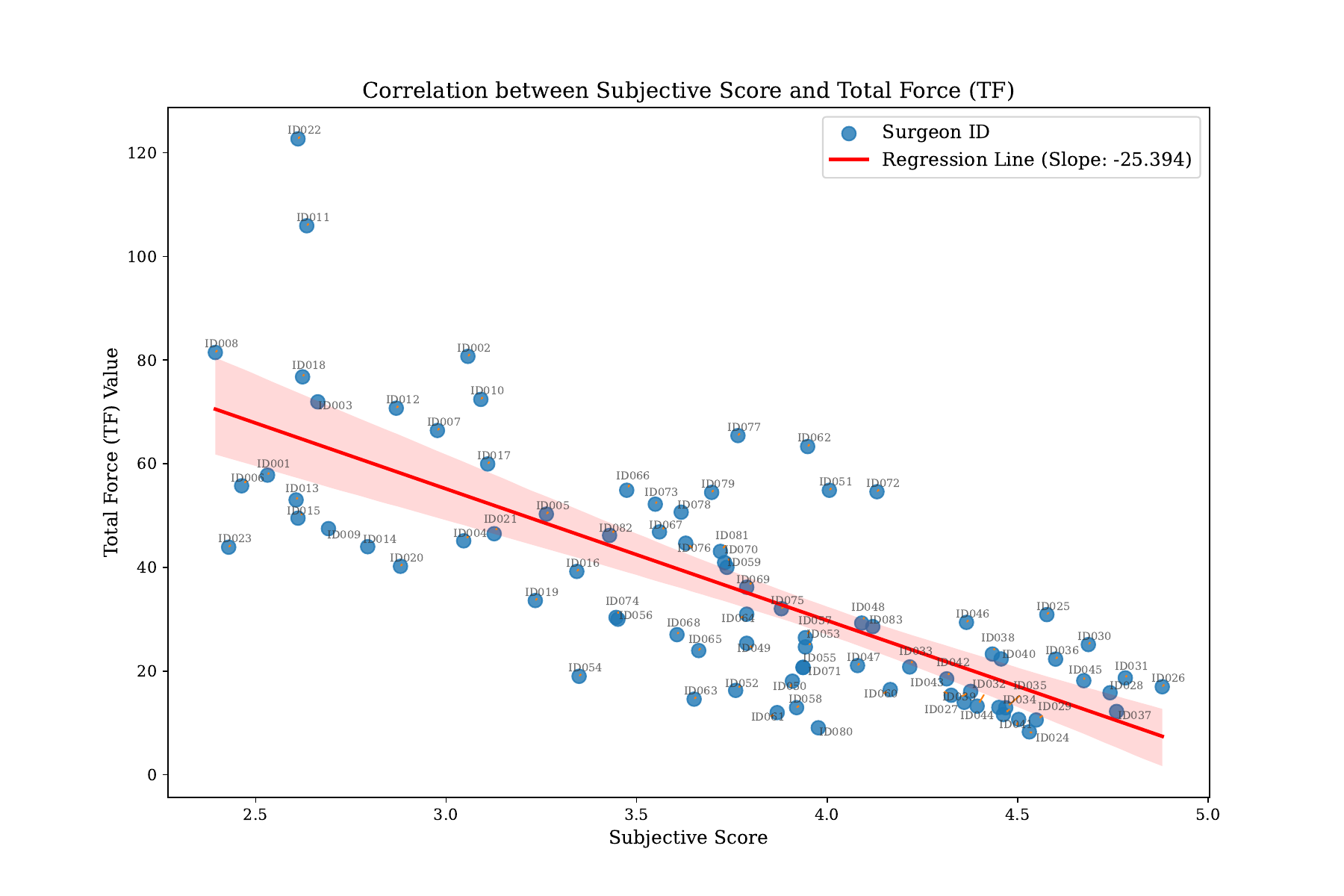}\label{TF}
    }
    \subfigure[Tissue Interaction (TI)]{
        \includegraphics[width=0.45\textwidth]{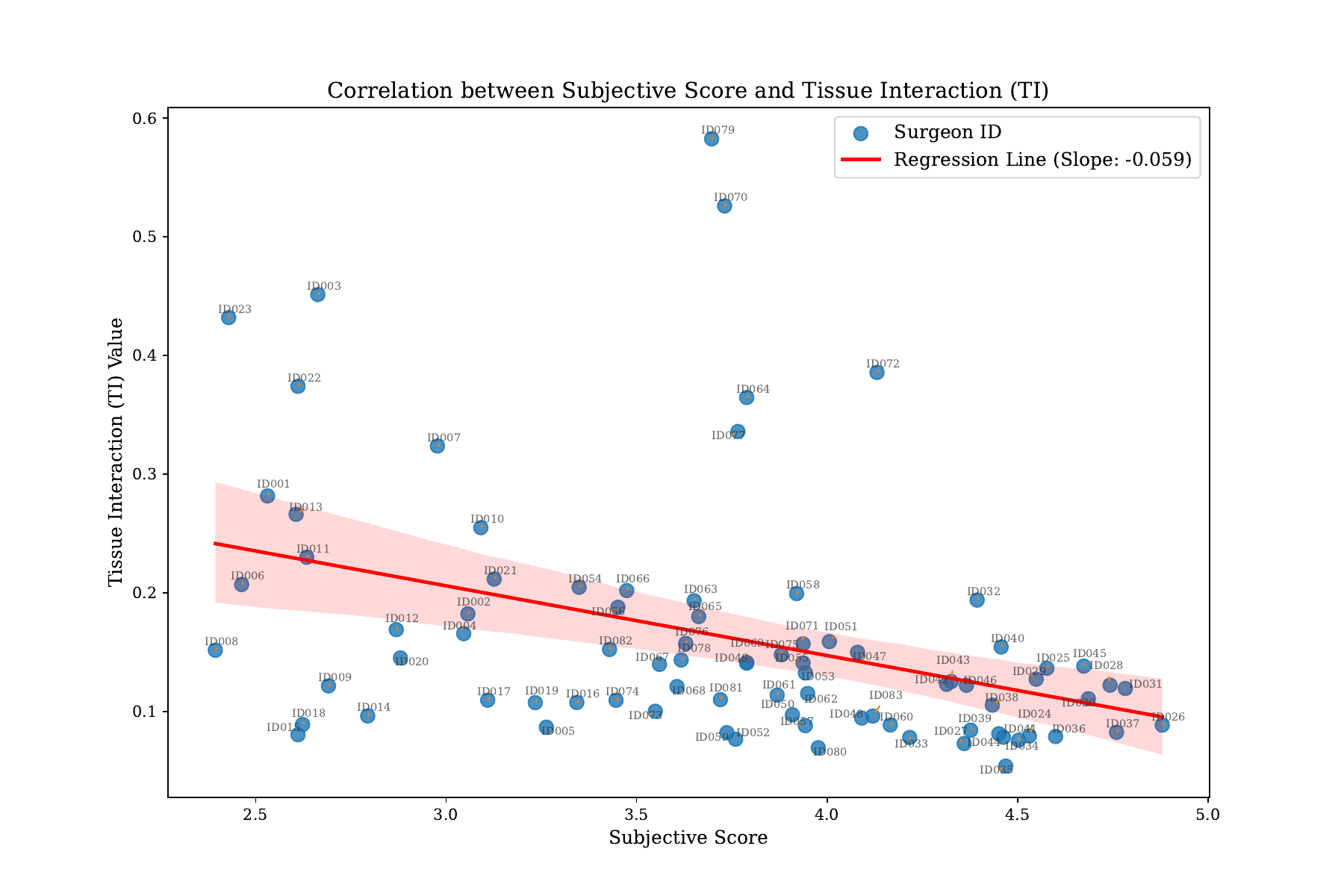}\label{TI}
    }

    \captionsetup{labelformat=empty}
    \caption{\textit{Figure~\ref{figcorrelation_plots}} continues on the next page.}
\end{figure}


\addtocounter{figure}{-1} 
\begin{figure}[htbp]
    \centering
    
    \subfigure[Speed Frequency Smoothness (SFS)]{
        \includegraphics[width=0.45\textwidth]{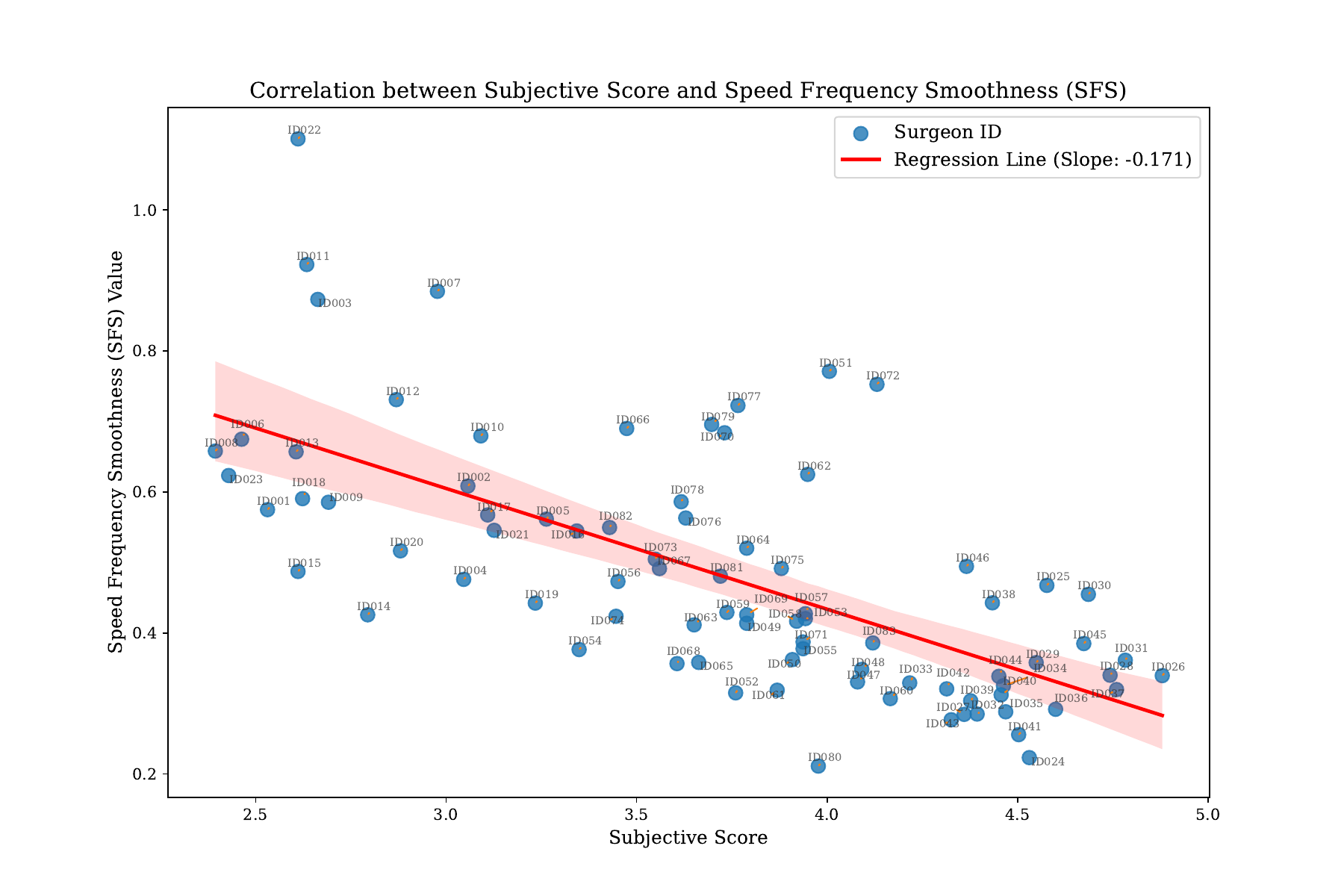}\label{SFS}
    }
    \subfigure[Spectral Entropy (SE)]{
        \includegraphics[width=0.45\textwidth]{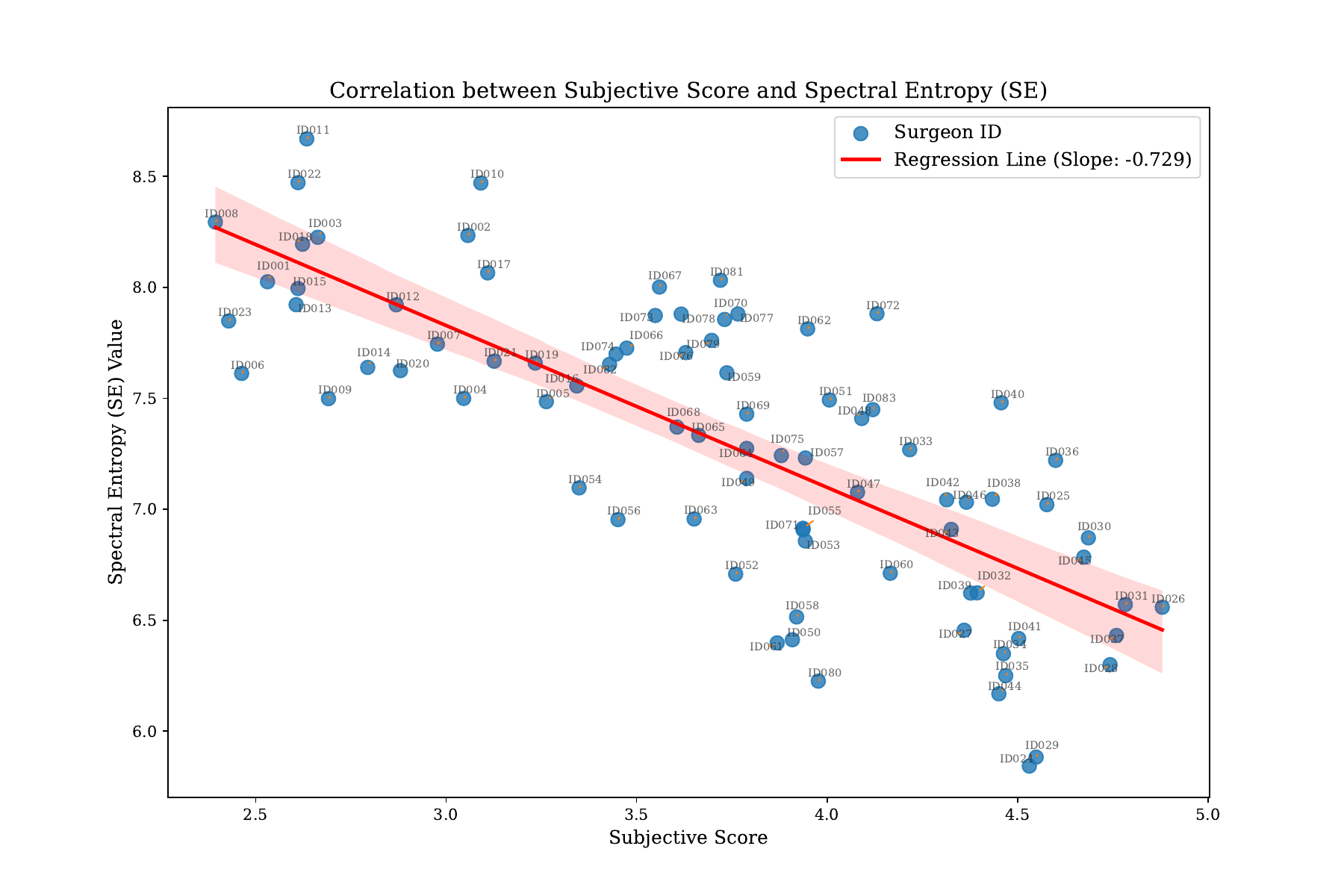}\label{SE}
    }

    \caption{Correlation plots for different metrics.}
    \label{figcorrelation_plots}
\end{figure}

\begin{enumerate}
    
\item 
Procedure Time (PT): As depicted in \textcolor{blue}{Figure} \ref{PT}, a significant inverse correlation was observed between Procedure Time (PT) and subjective skill scores, with a regression slope of –60.057. As skill scores increased, PT decreased, indicating that more proficient surgeons complete procedures more efficiently, likely due to enhanced technical dexterity and optimized decision-making. These findings highlight the value of PT as a metric for surgical competence, emphasizing its potential to improve patient outcomes and increase operating room efficiency.

\item 
Path Length (PL): Analysis of the PL metric underscores a negative regression slope of –272.946 in its correlation with subjective surgical skill scores (\textcolor{blue}{Figure} \ref{PL}). The shorter path lengths observed at higher skill levels suggest that experienced surgeons execute movements more efficiently, limiting unnecessary travel. This streamlined motion is particularly valuable during delicate procedures like capsulorhexis, where minimizing extra instrument movement enhances precision and patient outcomes.

\item 
Number of Alternating Movements (NAM): \textcolor{blue}{Figure} \ref{NAM} shows that the NAM exhibits a strong negative slope (–1840.795) when regressed against subjective surgical skill scores. Surgeons with higher proficiency perform fewer directional reversals, implying more controlled, deliberate motions and purposeful instrument handling. Reducing rapid back-and-forth movements enhances procedural efficiency and lessens the potential for tissue damage. These improvements in motor planning and execution are especially valuable in complex operations like capsulorhexis, where even minor inefficiencies can compromise patient outcomes.

\item 
Motion Difficulty (MD): In \textcolor{blue}{Figure} \ref{MD}, the analysis of the MD metric demonstrates a significant inverse relationship with subjective surgical skill scores, as indicated by a regression slope of –44.944. This result reveals that higher skill levels are associated with smoother movements, thereby minimizing abrupt changes in acceleration. This reduction in motion difficulty is critical for decreasing tissue trauma and achieving optimal surgical outcomes \cite{Hwang2005}.

\item 
Energy Consumption (EC): The analysis in \textcolor{blue}{Figure} \ref{EC} reveals a negative regression slope of –0.227, indicating that higher skill corresponds with lower energy consumption. This relationship suggests that proficient surgeons execute more efficient movements that minimize unnecessary kinetic energy expenditure. Consequently, such energy efficiency is a marker of refined motor control.

\item 
Number of Local Peaks (NLP): The regression analysis depicted in \textcolor{blue}{Figure} \ref{NLP} shows that the NLP is strongly negatively correlated with subjective surgical skill scores (slope: –1298.269). This finding suggests that more proficient surgeons achieve more fluid and controlled instrument movements, marked by fewer abrupt fluctuations. Enhanced motor control and precision, as evidenced by reduced NLP values, are vital in minimizing tissue trauma and improving surgical outcomes.

\item 
Total Force (TF): The analysis of the TF in \textcolor{blue}{Figure} \ref{TF} reveals a strong inverse correlation (slope: –25.394) with subjective skill scores. Surgeons with higher proficiency achieve more fluid and controlled instrument handling, reducing unnecessary force application.

\item 
Tissue Interaction (TI): The regression analysis in \textcolor{blue}{Figure} \ref{TI} shows a negative slope of –0.059, implying that increased surgical proficiency is associated with reduced TI values. Such reduction and consistency in motion reduce the likelihood of abrupt force fluctuations that could damage tissue, underscoring the importance of uniform and controlled acceleration patterns in ensuring successful outcomes in sensitive surgical interventions.

\item 
Speed Frequency Smoothness (SFS): In the frequency domain, abrupt and inadequately planned movements are typically represented by high-frequency coefficients, while smoother movements correspond to lower-frequency coefficients \cite{Soleymani2021}. The analysis depicted in \textcolor{blue}{Figure} \ref{SFS} shows that the SFS metric is inversely correlated with subjective surgical skill scores (regression slope: –0.171). Lower SFS values, associated with higher proficiency, indicate smoother speed profiles with fewer high-frequency fluctuations. 

\item 
Spectral Entropy (SE): The regression analysis in \textcolor{blue}{Figure} \ref{SE} shows that the spectral entropy (SE) metric exhibits a negative slope of –0.729. This result implies that higher-skill surgeons demonstrate lower SE values, meaning that the motion signal's energy is more tightly focused in particular frequency bands. This energy distribution pattern indicates consistent instrument movements and underscores improved motor control and precision.

\end{enumerate}

These correlations confirm the reliability of the proposed framework in replicating expert-based evaluations. By employing video-derived metrics as automated proxies for real-world competence, this approach enables consistent and scalable skill assessments across various surgical contexts.

In the final stage of our analysis, multiple clustering algorithms, including AggAverage, AggComplete, AggSingle, Agglomerative, BayesianGMM, Birch, BisectingKMeans, GaussianMixture, KMeans, MiniBatchKMeans, and Spectral, were employed to partition surgeons into two distinct groups based on their Overall Score (OS) as determined by expert subjective evaluation. The group exhibiting higher OS values was designated as “Expert (E),” whereas the lower-scoring group was labeled “Intermediate (I).” This OS-based clustering served as the ground truth for subsequent comparisons.

Subsequently, the same clustering techniques were applied independently to ten objective performance metrics derived from our intelligent assessment framework. Consistent with the inverse relationships observed in \autoref{figcorrelation_plots}, lower values of these metrics corresponded to the “Expert” cluster, while higher values mapped to the “Intermediate” cluster.

Subsequently, each objective metric’s clustering was compared against the OS-based clustering to assess how accurately it could distinguish between “Expert” and “Intermediate” surgeons. The formula below measures this accuracy by quantifying the overlap of surgeon IDs in the E and I clusters derived from each objective metric with the corresponding E and I clusters obtained via OS-based clustering. Thus, it provides a straightforward indicator of how well each objective metric aligns with the expert-derived groupings:

\begin{equation}
\text { Accuracy }=\frac{\left|\mathcal{E}_{\mathrm{OS}} \cap \mathcal{E}_{\text {metric }}\right|+\left|\mathcal{I}_{\mathrm{OS}} \cap \mathcal{I}_{\text {metric }}\right|}{\left|\mathcal{E}_{\mathrm{OS}}\right|+\left|\mathcal{I}_{\mathrm{OS}}\right|}
\end{equation}

where \(\mathcal{E}_{\text{OS}}\) and \(\mathcal{I}_{\text{OS}}\) are the sets of surgeons labeled “Expert” and “Intermediate” by the OS-based clustering, and \(\mathcal{E}_{\text{metric}}\) and \(\mathcal{I}_{\text{metric}}\) are the corresponding sets obtained from the metric-based clustering.

Table~\ref{tab:results} summarizes the highest accuracy achieved for each indicator. For each objective metric, the table lists the clustering method that produced the optimal accuracy when compared with OS-based labels, along with the corresponding threshold ranges (minimum and maximum values) for both the “Expert” and “Intermediate” clusters. Among all the objective indicators, NAM clustered via the Agglomerative method exhibited the greatest alignment with the surgeon groupings derived from subjective scoring.

In the case of the Overall Score, the table reports the aggregated thresholds across all methods, with an observed overlap in the range of 3.23 to 3.78. This overlap reflects the inherent variability in subjective assessments when consolidated over multiple clustering approaches.

\begin{table}[htbp]
    \centering
    \caption{Top clustering accuracies and threshold ranges for classifying surgeons into Expert and Intermediate groups.}
    \label{tab:results}
    \begin{tabular}{l l c l}
        \toprule
        \textbf{Indicator} & \textbf{Method} & \textbf{Accuracy (\%)} & \textbf{Thresholds for [E], [I]} \\
        \midrule
        PT & KMeans & 82 & [17-93], [97-288] \\
        PL & KMeans & 84 & [69-505], [544-1327] \\
        NAM & Agglomerative & \underline{\textbf{87}} & [404-3795], [4063-9135] \\
        MD & Spectral & 85 & [14-60], [63-217] \\
        EC & GaussianMixture & 82 & [0.06-0.39], [0.40-1.42] \\
        NLP & Spectral & 85 & [350-1582], [1616-6115] \\
        TF & Spectral & 85 & [8-34], [36-123] \\
        TI & GaussianMixture & 75 & [0.05-0.21], [0.22-0.58] \\
        SFS & Agglomerative & 81 & [0.21-0.52], [0.54-1.10] \\
        SE & Spectral & 84 & [5.84-7.37], [7.40-8.66] \\
        Overall Score (OS) & All Methods & - & [2.39-3.78], [3.23-4.88] \\
        \bottomrule
    \end{tabular}
\end{table}

Overall, our findings emphasize the critical role of precise hand-eye coordination and muscle control in surgical performance. These results align closely with previous research highlighting the importance of motion smoothness, controlled force application, and careful instrument handling for successful surgical outcomes \cite{Kletz2019,Nau2023,Aghazadeh2023,05838}.

Additionally, our study represents a significant advancement beyond prior investigations, such as Kim et al. \cite{Kim2019}, by examining a more comprehensive set of skill indicators derived from detailed motion data, coupled with a rigorously annotated dataset. Our analysis clearly demonstrates that expert surgical performance depends heavily on the effective coordination of visual perception, hand movements, and muscular precision.

\section{Conclusion}\label{sec3}

In this paper, we introduced an explainable AI framework designed to evaluate surgical skills in cataract procedures automatically. The objective framework presented here offers an accurate and scalable approach for assessing surgical proficiency. This approach has strong potential to complement or even replace existing subjective assessment systems like ICO-OSCAR. Furthermore, it can play a valuable role in refining surgical training programs by providing targeted, objective feedback, thereby systematically guiding trainees toward improved surgical outcomes.

The study stands out for several key achievements. First, we crafted the world’s largest repository of cataract surgery videos, encompassing 2,000 recordings, of which 83 were meticulously annotated using the newly established Capsulorhexis Skill Assessment System (CSAS). This dataset enabled a reliable benchmark for subjective (expert-driven) and objective (motion-based) measurements, thereby bridging a longstanding gap in AI-driven surgical education research. Notably, it surpasses JIGSAWS \cite{gao2014jigsaws} by capturing real surgical scenes rather than simulated tasks, and it improves upon Cataract-101 \cite{Cataract101} and Cataract-1K \cite{Ghamsarian2024} through a structured, multi-indicator annotation process conducted by three independent physicians.

Second, our approach proposed a data-driven pipeline that extracts clinically meaningful motion signals from the raw video stream. Through advanced computer vision algorithms for segmentation and tracking, we converted pixel-level motion data into quantifiable metrics that thoroughly reflect technical precision, smoothness, and tissue handling. This capability addresses the need for objective, reproducible indicators of surgical expertise.

Third, and most critically, we demonstrated that our ten computed motion-based metrics, ranging from path length to tissue interaction measures, exhibit robust correlations with expert-derived capsulorhexis skill ratings, achieving up to 87\% accuracy in skill classification. This level of performance indicates that automated, objective measures can match or even exceed traditional subjective evaluations, challenging the notion that high-quality surgical skill assessments must rely solely on human observers.

Fourth, our focus on explainability stands as a cornerstone achievement. By translating raw computational outputs into clinically interpretable indices, we offer not just an accuracy-driven tool but also a transparent framework readily understood by surgical educators and trainees. Such interpretability fosters trust in AI and provides targeted feedback, allowing surgeons to pinpoint, for example, whether excessive force, abrupt acceleration, or instrument misalignment underlies lower performance scores.

Despite these promising results, certain limitations persist. The current validation focuses heavily on the capsulorhexis phase, and additional studies are essential to ensure generalizability across other surgical tasks and clinical environments. The varying complexities of multi-institutional settings also underscore the necessity for broader validation. Furthermore, while the proposed metrics operate efficiently in retrospective analyses, the challenges of real-time feedback integration, such as processing speed, user interface design, and integration within the clinical workflow, remain outstanding.

Future directions for this research thus include (1) broadening the scope to capture other phases of cataract surgery (e.g., phacoemulsification and intraocular lens placement) and extending to different surgical specialties, (2) conducting large-scale, multi-center validation studies to account for diverse patient populations and healthcare infrastructures, (3) developing real-time coaching systems that combine motion analytics with intraoperative feedback to foster faster learning curves, and (4) incorporating advanced simulator and virtual reality platforms, allowing trainees to benefit from immediate, data-driven metrics in a safe, controlled setting, (5) undertaking a clinical study to assess and model the learning curve of surgical trainees employing this method, and (6) leveraging outputs from objective metrics and threshold-based clustering to automatically generate personalized skill reports, complete with strengths, weaknesses, and targeted suggestions, by integrating large language models (LLMs). Ultimately, such efforts may help alleviate global workforce shortages and democratize access to advanced surgical training by providing scalable, objective, and explainable AI-based tools.

\bibliography{sn-bibliography}

\end{document}